# Enhancing Active Learning for Sentinel 2 Imagery through Contrastive Learning and Uncertainty Estimation

D. Pogorzelski, P. Arlinghaus, and W. Zhang

*Abstract*— In this paper, we introduce a novel method designed to enhance label efficiency in satellite imagery analysis by integrating semi-supervised learning (SSL) with active learning strategies. Our approach utilizes contrastive learning together with uncertainty estimations via Monte Carlo Dropout (MC Dropout), with a particular focus on Sentinel-2 imagery analyzed using the Eurosat dataset. We explore the effectiveness of our method in scenarios featuring both balanced and unbalanced class distributions. Our results show that the proposed method performs better than several other popular methods in this field, enabling significant savings in labeling effort while maintaining high classification accuracy. These findings highlight the potential of our approach to facilitate scalable and cost-effective satellite image analysis, particularly advantageous for extensive environmental monitoring and land use classification tasks.

## I. INTRODUCTION

THE utilization of satellite imagery analysis has become an essential tool across diverse sectors, such as environmental monitoring, land-use and land-cover analysis, and disaster response [1][2][3]. With the continuous expansion in satellite data availability, there is a corresponding surge in the demand for sophisticated analytical techniques capable of efficiently processing this data and deriving actionable insights. Particularly, tasks such as land use classification and semantic segmentation of satellite images play pivotal roles in tracking and understanding temporal changes in landscapes [4]. Nevertheless, these critical tasks are often hindered by the requirement for extensive labeled datasets, which are both costly and labor-intensive to create [5].

Active learning is a strategic approach to mitigate the burden of labeling by strategically sampling a subset of the training data pool [6]. This subset ideally represents the broader dataset sufficiently to train effective models with minimal data. While active learning has a longstanding history, recent strides in deep learning within the realm of remote sensing have demonstrated a promising shift towards self-supervised learning (SSL) methods, which have shown considerable success in the active learning domain [7].

Despite the growing inclination towards using SSL for remote sensing applications, its integration within active learning strategies for remote sensing data remains underexplored. In this paper, we introduce a novel methodology to the existing repertoire of active learning frameworks, specifically tailored for the underdeveloped area of remote sensing active learning algorithms. This contribution aims to bridge the gap in the application of SSL in active learning, enhancing the efficiency and accuracy of remote sensing data analysis.

## II. RELATED WORK

*Active Learning*

Active learning is a subset selection methodology utilized in machine learning, particularly when labeled data is scarce or costly to obtain [6]. The fundamental goal of active learning is to strategically select an unlabeled subset $S_U$ from an unlabeled training set $D_U$ to train a model with a much lower amount of data, i.e. $|S_U| \ll |D_U|$, while aiming for a similar model performance. The cardinality or the sample size of $S$ is a hyperparameter. The selection is optimized to maximize the performance on a specific task $T$, e.g. classification or semantic segmentation. Let $S_U^*$ denote the set of all subsets of $D_U$ with size $N$ each. Mathematically, the objective of active learning can then be expressed as

$$S_U' = \underset{S_U \in S_U^*}{\arg\max} \; J(T)$$

where $J$ describes the performance. The strategy of maximizing the function is called *query strategy* and the effectiveness of $S'_U$ is typically evaluated through metrics that measure model performance, such as accuracy, precision, or a domain-specific evaluation criterion, reflecting the active learning cycle's contribution towards achieving more informative and representative training samples. In a neural network setting, the performance is usually measured by a loss function. Given a

This study is a contribution to the Helmholtz Imaging Platform project "AutoCoast - Automatic detection of coastline change and causal linkage with natural and human drivers" (ZT-I-PF-4-048). It is also supported by the Helmholtz PoF programme "The Changing Earth – Sustaining our Future" on its Topic 4: Coastal zones at a time of global change.

David Pogorzelski, Peter Arlinghaus and Wenyan Zhang are with the Department of Sediment Transport and Morphodynamics, Helmholtz-Zentrum Hereon, 21502 Geesthacht, Germany (e-mail: david.pogorzelski@hereon.de; peter.arlinghaus@hereon.de; wenyan.zhang@hereon.de)

neural network $f(x)$ and a corresponding loss function $L(x)$, the active learning objective can be expressed as

$$S_U' = \underset{S_U \in S_U'}{\operatorname{argmin}} L\big(f(x)\big)$$

The process can be repeated which makes it an *iterative active learning* scheme:

$$S_U^{(i+1)} = \underset{S^{(i)} \in S_U^{*(i)}, \ S_U^{(i)} \cap S_L^{(i)} = \emptyset}{\operatorname{argmin}} L\left(f_{S_L^{(i)}}(x)\right)$$

$$S_L^{(i+1)} = S_L^{(i)} \cup a\big(S_U^{(i+1)}\big)$$

$$S_U^{*(i+1)} = S_U^{*(i)} \setminus S_L^{(i+1)}$$

where $f_{S_{L^{(i)}}}$ describes the model learned on the labeled data $S_L$, a(x) denotes the mapping from unlabeled to labeled data. The process is repeated for all $i$ until a convergence criterion is met, e.g. a predefined accuracy. The advantage of iterative active learning is the continual enhancement of selection decisions with each iteration.

*Contrastive Learning*

A common technique in the training of deep learning models is the combination of pre-training and fine-tuning, together called transfer learning [8].

Pre-training is the process of training a neural network on a large, pre-processed dataset to develop a foundational model that can be adapted for various tasks. This foundational model captures general features that are useful across different domains. By transferring the learned parameters from the pre-trained model to a new model, one can leverage these pre-existing insights. This method is especially useful as it allows the new model to start from an advanced point of learning. The next step involves fine-tuning, where the pre-trained model is further trained on a specific dataset of interest, allowing it to specialize and adapt to the nuances of that data. This approach significantly accelerates the training process and improves the model's performance on specialized tasks. Often, a small amount of data is already enough to fine-tune a pre-trained model [7].

This approach is well-established in domains like image processing, which benefit from the availability of large datasets. However, in fields where labeled data is scarce, traditional supervised learning approaches are not viable. In such cases, alternative methods such as *self-supervised learning* (SSL) are utilized. SSL leverages unlabeled data to learn feature representations that are useful for downstream tasks. This method enables the extraction of meaningful patterns from data without the need for labels. A broad overview of SSL in remote sensing is given in [7].

SSL can be used for different objectives like reconstructing data, predicting self-produced labels, or learning a representation that maps semantically similar inputs close together in the representation/ feature space $f$ such that $|f(x_1) - f(x_2)| \to 0$.

In this paper, we focus on the latter part, also referred to as *contrastive learning*.

To map semantically similar inputs together, the model could trivially map all instances to one common point where the similarity task would be fulfilled. However, this would lead to a poor representation since also dissimilar points would share similar representations. This is called *model collapsing*. One way to prevent model collapsing, is by the means of *negative sampling* which incorporates negative, dissimilar samples as well:

$$sim\big(f(x), f(x^+)\big) \gg sim\big(f(x), f(x^-)\big)$$

where $x$ is referred to as the *anchor point* and $x^+$ as the positive sample and $x^-$ as the negative sample. $f$ is the encoding function, usually a neural network, and $sim$ describes a similarity function where a larger value describes higher similarity between two objects.

There are many ways to generate positive and negative samples for an anchor point and to incorporate these in a training framework.

A common way is to augment the anchor point by transformations like rotation, cropping etc. Each augmented version is also referred to as a *view* of an anchor point.

A prominent loss function for negative sampling contrastive learning is the NT-Xent (Normalized Temperature-scaled Cross Entropy) loss:

$$L(i,j) = -\log\left(\frac{\exp\big(sim(z_i, z_j)/\tau\big)}{\sum_{k=1}^{2N} \mathbb{1}_{k \neq i} \exp(sim(z_i, z_k)/\tau)}\right)$$

Given a data instance $x$ from the training set D with size $|D| = N$, $z_i$ and $z_j$ are two different, but similar views of $x$. $z_k$ includes all other samples plus $z_j$, the positive pair. The number of summands is $2N$ since two views are generated for each instance of $x$. $z_i$ is also referred to as the query, $z_j$ and $z_k$ the key respectively. $\tau$ is a hyperparameter and controls the decision boundary between positive and negative samples. A large $\tau$ leads to a softer decision boundary and vice versa. The indicator function $\mathbb{1}$ ensures that the query itself is not included in the comparison.

A popular SSL framework is MoCo [9]. [7] showed that MoCo generally outperforms other well-known SSL methods in remote sensing classification tasks. Additionally, it was shown that models fine-tuned with 50% labeled data can achieve similar performance to those trained with full supervision. A crucial finding is the significant role of random cropping as a data augmentation strategy in enhancing the effectiveness of the encoder. The Eurosat dataset was specifically highlighted to demonstrate the advantages of SSL in analyzing remote sensing data.

*Uncertainty Estimation*

Another query strategy in active learning involves the computation of uncertainty. In the Bayesian framework, uncertainty is typically represented by the posterior distribution $p(\theta|D)$, where $\theta$ denotes model parameters and $D$ the observed data. However, the computation of this posterior distribution becomes computationally infeasible in deep learning due to the extremely high dimensionality of the parameter space. A practical and widely acknowledged solution to this challenge is Monte Carlo (MC) Dropout [10]. MC Dropout not only simplifies the estimation of the posterior but also provides a computationally efficient approximation by utilizing dropout at both training and inference stages. This technique enables the model to generate different outputs for the same input by randomly dropping units, which effectively samples from an approximate posterior distribution.

It was applied in another work for active learning in image classification [11]. The results showed that uncertainty estimation helps reducing the labelling effort.

*Combined Methods*

Another study presented in [12] explores a hybrid approach for active learning in remote sensing. Initially, contrastive learning with negative sampling was employed using MoCo to encode the data, followed by clustering of the encoded data. Subsequently, Euclidean-distance-based diversity sampling was applied within each cluster to select training samples for the main model. This one-shot procedure demonstrated superior performance over other competitive methods.

In our approach, termed MCFPS, we integrate the beneficial aspects of the methods described above. We utilize negative sample contrastive learning with MoCo to encode data into a lower-dimensional feature space. Subsequently, we explore the neighborhood of each sample to identify and select the most uncertain sample for training our main model.

.

### III. METHODOLOGY

Our proposed method consists of an initialization and an iterative learning scheme:

*Initialization*

Given a dataset $X \in \mathbb{R}^{N \times d}$, we start by training an encoder $f$ to transform the dataset to $X_{enc} \in \mathbb{R}^{N \times e}$, where $e \ll d$.

*Iterative Learning Scheme*

The iterative learning scheme consists of five steps:

**1. Model initialization:** With the encoded data $X_{enc}$, we initiate our iterative active learning scheme. This involves initializing a model $g$ to estimate uncertainty using MC dropout.

**2. Diversity Sampling:** We employ farthest-point-sampling within the transformed space $X_{enc}$ to select $S$ objects from the dataset.

**3. Nearest-Neighbor Search:** For each sampled object $S_i$, a nearest-neighbor search is conducted with a neighborhood size of $k$.

**4. Uncertainty Estimation:** Uncertainty for each $S_i$ within its neighborhood $M_i$ is estimated through $t$ forward passes using model $g$.

**5. Candidate Selection:** For classification tasks with $C$ classes the previous step yields the uncertainty $\hat{Y} \in \mathbb{R}^{t \times C}$ for each neighbor in $M_i$. The overall uncertainty is determined by averaging the outcomes to yield $\hat{Y}_{Mean} \in \mathbb{R}^{C}$ and taking the maximum of $\hat{Y}_{Mean}$

The final step involves selecting the sample with the highest uncertainty in each neighborhood for inclusion in the set of samples that need to be labeled by a human annotator next. The labeled data is then used to pre-train the model $g$ that is being used for uncertainty estimation for the next iteration.

Additional logic can be added, such as skipping samples for which $g$ outputs a certainty higher than a predefined threshold.

### IV. EXPERIMENTATION AND RESULTS

*Experimental Setup*

The code for our experiments can be found here: https://github.com/autocoast/active-learning-sentinel-s2

For our experiments we used the Eurosat dataset [13], a popular remote sensing dataset with 27000 Sentinel S2 image patches each of size 64x64x13. Every patch is assigned to one out of 10 possible classes. The dataset's class distribution is well balanced.

A pre-trained ResNet50 model, sourced from [14], served as our encoder. The model is the result of an extensive MoCo-based training on the BigEearthNet dataset [14][15]. The encoder transforms each 64x64x13 patch into a 2048-dimensional vector. Figure 1 illustrates the encoded data after an additional t-SNE transformation with two components. The different classes are distinguished by varying colors, demonstrating the effective separability of the data. To quantify the impact of this method, we applied a k-Nearest-Neighbor (kNN) classifier to the t-SNE-reduced data, achieving a test accuracy of 95%. Conversely, direct encoding of the raw 64x64x13 data, subsequent projection onto two t-SNE components and kNN classification with the same train/test split, resulted in a test accuracy of only 68%.

Our experimental framework was executed under two scenarios. Initially, we trained our models on the original, class-balanced dataset. In the second scenario, we introduced imbalances by randomly removing samples from each class, creating four variations of the dataset. Each dataset, both balanced and each variation of the unbalanced dataset, was split in the same way: 80% for training and 20% for testing. Details on the specific parameters used, including the seeds for randomization, are available in the provided GitHub repository.

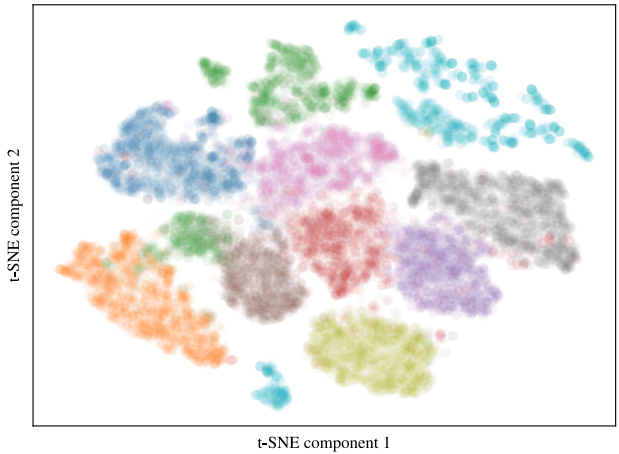

Fig. 1. T-SNE plot of the Eurosat dataset after performing the encoding with SSL and further PCA employment. Each color represents a label in the dataset.

For each dataset configuration—both balanced and the four unbalanced datasets—we compared our method (MCFPS) against three alternatives: random selection, farthest point sampling (FPS) alone, and a one-shot active learning (OSAL) approach similar to [12]. For FPS, our learning algorithm proceeded only up to step 2 of our method in each iteration. For OSAL, we followed the steps recommended in [12], clustering our encoded space (2048 dimensions) using k-means++ with the number of clusters determined by the silhouette score, which resulted in 2 clusters. Within each cluster, farthest point sampling was applied to select the candidates. Although OSAL is a one-shot approach, we simulated an iterative process by performing one-shot for each iteration, using the candidates selected in previous iterations as the initial candidates. For each iteration, we skipped samples with a certainty estimation exceeding 80% for the MCFPS case.

To manage computation time, we limited each method to eight iterations for both the balanced and unbalanced scenarios, selecting 64 candidates in each round. Additionally, achieving a test accuracy of 90% was established as a secondary stopping criterion. Thus, a maximum of 512 candidates were sampled per method unless the active learning process stopped earlier.

Before comparing the three methods, we first validate the effectiveness of a pre-trained ResNet-50 model using self-supervised learning (SSL) as demonstrated in [7]. To achieve this, we evaluate the test accuracies on the Eurosat dataset with three different ResNet-50 models: a non-pre-trained model, an ImageNetV2-based pre-trained model, and the SSL-pre-trained model. The comparison is conducted using random subsets of 2%, 5%, and 10% of the data in a balanced setting, with an 80/20 training/test split as previously described.

*Results*

Figure 2 demonstrates the effectiveness of a pre-trained model utilizing SSL. Notably, the model achieves an accuracy of approximately 90% with just 2% of the data. As the available training data increases to 5% and 10%, the SSL-pre-trained model continues to outperform, reaching test accuracies of over

94% and 95%, respectively. In practical terms, this indicates a significant reduction in the number of labeled images required to achieve high accuracy, highlighting the benefits of SSL pre-training in scenarios with limited labeled data.

Figures 3 and 4 present the results when applying FPS, OSAL, and MCFPS as query strategies. In the balanced dataset scenario, random selection requires fewer labels to achieve 90% test accuracy compared to FPS and MCFPS. However, MCFPS outperforms FPS alone. OSAL shows a similar increase in accuracy in the first two iterations but suddenly stops improving and converges at roughly 75% accuracy.

For the unbalanced dataset, both FPS and MCFPS outperform random selection. MCFPS yields slightly better results than FPS. Furthermore, no random selection run could achieve the 90% accuracy mark within the 8 iterations criterion. OSAL falls significantly below all other methods and shows a wide spread of confidence with generally low accuracy. While random, FPS and MCFPS show a steep increase in accuracy during the initial iterations, FPS and MCFPS demonstrate a faster improvement rate with each subsequent iteration. MCFPS shows a quicker and more stable convergence to higher accuracy levels, reaching the target accuracy of 90% slightly faster and with greater stability compared to FPS.

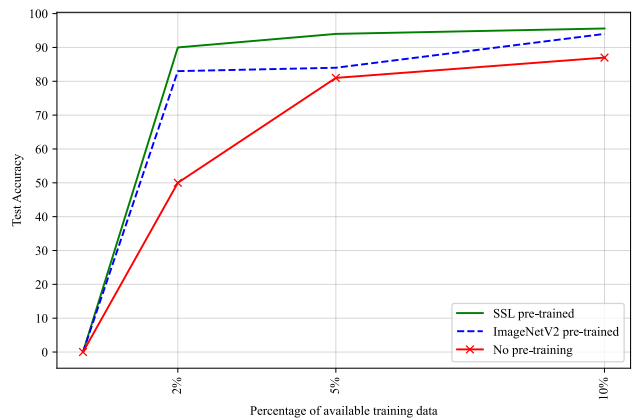

Fig. 2. Utilizing a contrastive learning based pre-trained model significantly boosts the performance of the model. This was already shown in [7].

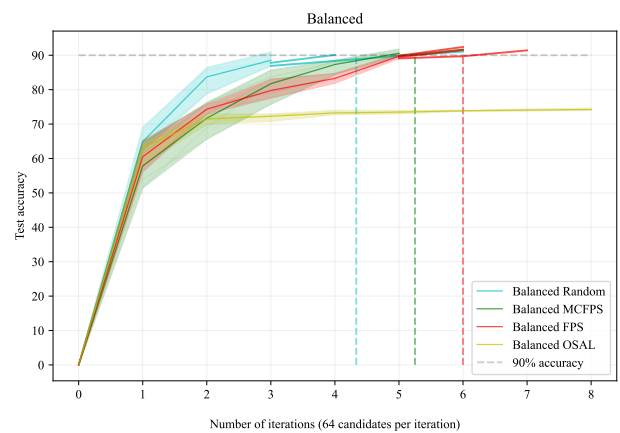

Fig. 3. In the balanced scenario, the random selection method outperforms both the FPS and MCFPS methods. Additionally, the MCFPS method

demonstrates superior performance compared to the FPS method. OSAL falls significantly below the results of the other methods. The colored dashed lines indicate the average values for all methods reaching 90% accuracy. The shaded area summarizes all runs with mean and spread. The area ends where some runs already reached 90%. After that, single lines are used to mark each individual run.

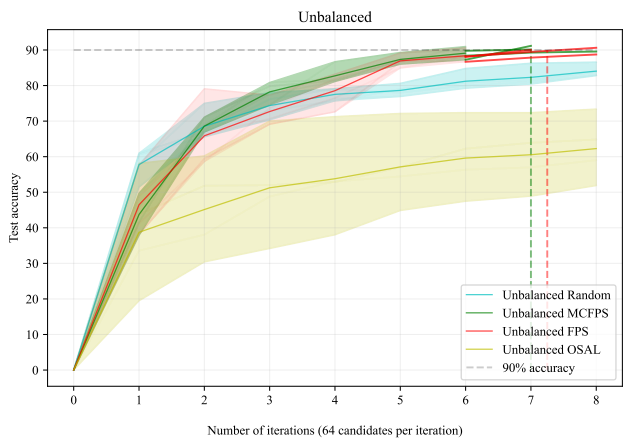

Fig. 4. In the unbalanced setting, random selection performs worse compared to FPS and MCFPS. However, as previously observed, MCFPS slightly outperforms pure FPS. OSAL falls significantly below the results of the other methods. The colored dashed lines indicate the average values for all methods reaching 90% accuracy, with no random run achieving this mark. The shaded area summarizes all runs with mean and spread. The area ends where some runs already reached 90%. After that, single lines are used to mark each individual run.

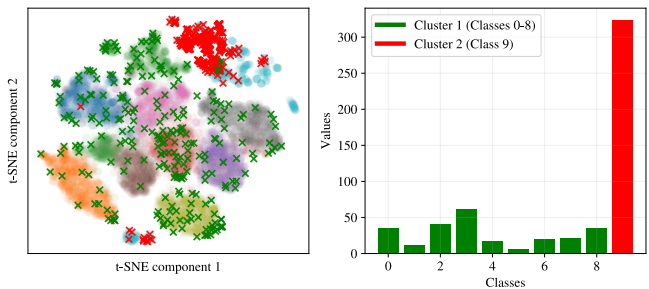

Fig. 5. The OSAL query strategy for two clusters. It can clearly be seen that selecting candidates solely based on silhouette score-based clustering and farthest point sampling alone is not sufficient to properly select candidates. The left plot shows the t-SNE visualization of the cluster result obtained, with Cluster 1 (Classes 0-8) in green and Cluster 2 (Class 9) in red. The right bar plot displays the distribution of values for each class, highlighting the significant imbalance between Cluster 1 and Cluster 2.

## V. DISCUSSION

The results indicate that utilizing a properly SSL pre-trained model significantly reduces the necessary labeling effort. To further enhance this reduction, it is essential to differentiate between class-balanced and class-unbalanced cases.

In class-balanced scenarios, random sampling appears to be sufficient. However, in class-unbalanced situations, which are more common in real-world applications - see for example [5] - diversity and uncertainty-based active learning methods prove to be particularly beneficial. MCFPS shows slightly better performance than FPS in both scenarios. The poor performance of OSAL can be attributed to the clustering result based on the silhouette score. As depicted in Figure 5, one can clearly see the class imbalance. This occurs because candidates are picked equally from each cluster in each iteration. Additional logic might help to overcome this issue.

Our framework includes several tunable hyperparameters, such as neighborhood size, the number of dropout layers, the number of forward passes for uncertainty measurements, and embedding dimensions. Optimizing these parameters could potentially lead to even better results.

Although our focus is on Sentinel-2 data, we believe that the method is applicable to other instruments and even other domains. This flexibility suggests that the approach could be beneficial across various types of remote sensing data and potentially other fields that require efficient labeling strategies.

In conclusion, our findings demonstrate the potential of SSL pre-trained models combined with diversity and uncertainty-based active learning to minimize labeling efforts. They also highlight the importance of tailored active learning strategies for different data scenarios. Future work will focus on extending these methods to more challenging tasks and further optimizing the framework to achieve superior performance.